\documentclass{article}
\usepackage{cite}
\usepackage{amsmath,amssymb,amsfonts}
\usepackage{tablefootnote}
\usepackage{threeparttable}
\usepackage{tabularray}
\usepackage{array}
\usepackage{multirow}
\usepackage{algorithmic}
\usepackage{graphicx}
\usepackage{textcomp}
\usepackage{comment}
\usepackage{hyperref}
\usepackage{geometry}
\title{Spatial Temporal Approach for High-Resolution Gridded Wind Forecasting across Southwest Western Australia}
\author{Fuling Chen\thanks{International Centre for Radio Astronomy Research, University of Western Australia, WA 6009 Australia. Email: fuling.chen@uwa.edu.au}
\and
Kevin Vinsen\thanks{International Centre for Radio Astronomy Research, University of Western Australia, WA 6009 Australia.}
\and
Arthur Filoche\thanks{International Centre for Radio Astronomy Research, University of Western Australia, WA 6009 Australia. School of Earth Sciences, UWA Oceans Institute, University of Western Australia, WA 6009 Australia.}
}

\date{Date of publication: xxxx 00, 0000 \\
      Date of current version: xxxx 00, 0000}

\begin{document}

\maketitle

\begin{abstract}
Accurate wind speed and direction forecasting is paramount across many sectors, spanning agriculture, renewable energy generation, and bushfire management. 
However, conventional forecasting models encounter significant challenges in precisely predicting wind conditions at high spatial resolutions for individual locations or small geographical areas (< 20 $\text{km}^2$) and capturing medium to long-range temporal trends and comprehensive spatio-temporal patterns.  
This study focuses on a spatial temporal approach for high-resolution gridded wind forecasting at the height of 3 and 10 metres across large areas of the Southwest of Western Australia to overcome these challenges. 
The model utilises the data that covers a broad geographic area and harnesses a diverse array of meteorological factors, including terrain characteristics, air pressure, 10-metre wind forecasts from the European Centre for Medium-Range Weather Forecasts, and limited observation data from sparsely distributed weather stations (such as 3-metre wind profiles, humidity, and temperature), the model demonstrates promising advancements in wind forecasting accuracy and reliability across the entire region of interest. 
This paper shows the potential of our machine learning model for wind forecasts across various prediction horizons and spatial coverage. 
It can help facilitate more informed decision-making and enhance resilience across critical sectors.
\end{abstract}

\maketitle

\section{Introduction}
\label{sec:introduction}
\subsection{Background}
Accurate forecasting of wind speed and direction is paramount across various domains, playing a pivotal role in weather prediction, renewable energy generation, agricultural management, and bushfire mitigation efforts. 
Accurate predictions enable meteorologists to deepen their understanding of atmospheric processes, leading to more precise weather forecasts and timely alerts for severe weather events \cite{stumpf2021national}. In the realm of renewable energy, precise forecasts of wind conditions are indispensable to optimise the performance of wind farms and integrate wind energy efficiently into the power grid \cite{soman2010review,wang2011review,santhosh2020current}. 
In agriculture, wind forecasts inform critical decisions such as crop spraying, sprinkler or central pivot irrigation timing, and pest control, ultimately improving crop yields and water management \cite{pujahari2022intelligent}. 
For bush-fire management, timely and accurate predictions of wind speed and direction are crucial for modelling fire behaviour, planning firefighter deployment, and planning evacuations, thereby reducing the impact of bushfires on communities and ecosystems \cite{taylor2013wildfire,coen2020computational}. 
Given the multifaceted applications of wind forecasting, advancements in machine learning-based techniques for predicting wind speed and direction hold immense promise for bolstering societal resilience and fostering sustainable development.

Traditionally, wind forecasting models fall into three categories: physical, statistical time series analysis and machine learning. Physical models, such as numerical weather prediction (NWP) and weather research forecasting (WRF), provide insights into weather patterns and long-term climate trends \cite{al2010review,storm2009evaluation}, but may lack accuracy for short-term, localised forecasts, especially in complex terrains. Statistical methods, including the autoregressive moving average (ARMA) and autoregressive integrated moving average (ARIMA) \cite{rajagopalan2009wind,abdelaziz2012short,yuan2017wind,maatallah2015recursive}, leverage historical data for wind speed prediction but may be limited by their reliance on linear relationships. 
Artificial intelligence techniques, like machine learning and neural networks, have revolutionised wind forecasting, offering models such as support vector machines (SVM), long short-term memory (LSTM), and convolutional neural networks (CNN) \cite{akhtar2021average,qu2019multi,kumar2016generalized,shahid2021novel,solas2019convolutional,joseph2023near,li2019short,mezaache2018auto}, which excel in short-term forecasting but face challenges in capturing spatial and temporal characteristics of wind, as well as the ability in medium to long term forecasting.

We can categorise forecasting models into four distinct prediction horizons: very short-term (spanning seconds to 30 minutes ahead), short-term (ranging from 30 minutes to 6 hours ahead), medium-term (encompassing 6 hours to 1 day ahead), and long-term (extending from 1 day to 10 days or more ahead) \cite{wang2011review}.
While most localised models focus on very short-term to short-term predictions due to their emphasis on single sites and immediate practical applications; advancements in artificial intelligence techniques are extending the forecast horizon to capture medium and long-term trends. 
Global forecasts tend to focus primarily on medium to long-term forecasts.
However, challenges persist in accurately predicting wind conditions beyond short-term horizons, including model complexity and data availability constraints.

Since wind represents airflow propagating rapidly across large-scale areas, the wind speed and direction at one site will exhibit strong spatial and temporal correlations with nearby sites \cite{ren2020spatial}. 
Extensive research, as shown in Table~\ref{tab:prevstudy}, highlights the significant impact of these correlations on wind prediction. Various models have been developed to harness these characteristics and forecast wind conditions for broader regions. Table~\ref{tab:prevstudy} shows studies that have focused on building spatial and temporal models for predicting wind speed across various sites in a local area. Interestingly, we found no prior applications of machine learning models for wind speed and direction prediction in a gridded area.
These models utilised the historical data, typically just the wind speed at a specific height, to forecast future conditions. 
However, in many cases, obtaining ground truth data for all stations across an area can be challenging, limiting the accuracy and reliability of predictions.

\begin{table}
\centering
\caption{\textbf{Previous studies using spatio-temporal models predicting wind in an area} \label{tab:prevstudy}}
\label{tab:prevstudy}
\begin{threeparttable}
\begin{tblr}{
  vline{2-4} = {-}{},
  hline{2} = {-}{},
}
Ref & Area* & {Prediction \\Horizon} & Data\\
\cite{zheng2021spatio} & 4 $\times$ 4 wind farms & 15 m - 3 h & wind speed\\
\cite{zhu2018wind} & 10 $\times$ 10 wind farms & 5 - 60 m & wind speed\\
\cite{khodayar2018spatio} & 145 wind farms & 1 - 24 h & {wind speed \\and direction}\\
\cite{hu2019very}& 28 wind farms & 30 m & wind speed\\
\cite{zhu2019learning} & 10 $\times$ 10 wind farms & 3 h & wind speed\\
\cite{pei2022short} & 20 wind farms & 15 m & wind speed\\
\cite{christoforou2023spatio} & {5 wind farms across\\ 3.64$^\circ$ $\times$ 2.39$^\circ$}& 6 h & wind speed\\
\cite{liu2020probabilistic} & {case 1: 421 wind farms \\across 0.6$^{\circ}$ $\times$ 0.34$^{\circ}$ \\~\text{case 2}: 20 $\times$ 20 wind farms\\ across 10$^{\circ}$ $\times$ 10$^{\circ}$} & 6 h & {wind speed \\and direction, \\temperature, \\dew point, \\humidity, \\rainfall}
\end{tblr}
\begin{tablenotes}
\footnotesize
\item[*] The typical spacing between wind turbines in a wind farm ranges from 8 to 12 times the rotor diameter, with an average rotor diameter exceeding 130 metres as of 2022. This translates to a practical spacing of approximately 1.08 km to 1.56 km between turbines.
\end{tablenotes}
\end{threeparttable}
\end{table}

Forecasting wind speed and direction at a height of 10 metres entails consideration of various geographical and meteorological features. 
Wind at a lower level, typically measured at 3 metres, is a crucial input due to its direct influence on wind patterns and behaviour closer to the ground \cite{maklad2014generation}. 
The terrain features play a significant role, as topographical variations can cause wind acceleration or deceleration, creating local wind patterns and turbulence \cite{jimenez2011surface}. 
Mean sea level air pressure \cite{miller2003once,ducet1999response} and temperature gradients \cite{chang2017oscillation} also contribute to wind dynamics, influencing the atmospheric circulation patterns that drive wind movement. 
These factors collectively interact to shape wind characteristics at the specified height, necessitating their inclusion in predictive models to accurately forecast wind behaviour and variability.

\subsection{Motivations and Contributions}
The present study is motivated by the observation that existing models primarily focus on short-term and very short-term predictions (see column ``Prediction Horizon'' in Table \ref{tab:prevstudy}), yet there is a pressing need for medium-term forecasts. 
Medium-term forecasts are indispensable across sectors like agriculture, wind energy generation, and bushfire management. 

The second motivation arises from a common trend in spatial temporal modelling: previous studies have typically concentrated on either small, densely predicted areas like wind farm arrays or sparsely distributed locations across larger areas with low resolution (shown in the column ``Area'' in Table \ref{tab:prevstudy}). 
However, this approach presents challenges in capturing comprehensive spatial and temporal patterns, hampering the models' effectiveness in longer-term predictions and broader geographic coverage. 
As a result, existing studies have been limited in their ability to provide accurate forecasts for extended time horizons and over larger geographical extents.

The third motivation stems from a noteworthy observation in previous machine learning studies. 
While historical wind speed, direction, terrain, air pressure, temperature, and humidity have been identified as crucial factors in wind forecasting, a significant disparity exists in their utilisation across other research projects (see the column of ``Data'' in Table \ref{tab:prevstudy}). 
Specifically, most studies have focused on historical wind speed, with only a limited number incorporating wind direction into their analyses. 
Integrating temperature, humidity data, and terrain information has been even rarer, with just a solitary study \cite{liu2020probabilistic} identified in the literature utilising temperature and humidity data. To the authors' knowledge, among the studies outlined in Table \ref{tab:prevstudy}, and in conjunction with research focusing on wind forecasting in complex terrain \cite{qiao2022wind,acikgoz2020extreme}, only four of the referenced meteorological studies have incorporated terrain data into their machine learning models \cite{le2023emulating,dujardin2022wind,martin2021fine,milla2024economical}. Recognising that deep neural networks allow the use of contextual information more naturally than classical statistical models, our study seeks to bridge this gap by amalgamating advanced machine-learning techniques with a comprehensive suite of meteorological factors, including mean sea level air pressure, humidity, and temperature; and terrain data. 
Our research endeavours to demonstrate the efficacy of this holistic approach, showcasing promising performance in wind forecasting accuracy and reliability.

\section{The proposed model}
\label{sec:approach}
We applied the attention-based encoder-decoder (ABED) architecture, illustrated in Appendix \ref{sec:app1} Figure \ref{fig:model} , designed for forecasting wind speed and direction at the height of 10 metres. 
Initially developed for tasks such as neural machine translation, image caption generation, video description generation, and end-to-end neural speech recognition \cite{cho2015describing}, ABED has since found applications in multivariate time series forecasting \cite{du2020multivariate} and sea surface height interpolation \cite{archambault2023unsupervised}.

Our input data comprises fully gridded data and sparse observation data, amalgamated into a 5-dimensional array ($n_b \times n_f \times n_t \times n_{lat} \times n_{lon}$), where $n_b$, $n_f$, $n_t$, $n_{lat}$, $n_{lon}$ represent the batch size, number of features, time length, latitude bins, and longitude bins, respectively. 
Further details on the input data are provided in Section \ref{sec:case:datadescrip}.

The encoder commences with batch normalisation, followed by a 3D convolution in both time and spatial dimensions, resulting in 4 output channels representing feature patterns. 
Subsequent downsampling blocks progressively reduce spatial dimensions by a factor of 2 while enriching channels to 16 (see Fig \ref{fig:model}).

The decoder comprises stacked deep residual sequence and spatial attention blocks (N-RSSAB), upsampling blocks, and a final 3D convolution. 
Inspired by previous work \cite{che2022ed, woo2018cbam}, N-RSSAB constructs a trainable decoder network to learn sequence-wise and spatial-wise features adaptively. 
Unlike previous methods that employ global skip connections \cite{che2022ed}, N-RSSAB is sequentially incorporated into the decoder to capture local transient patterns progressively. 
Each RSSAB within N-RSSAB starts with two 3D convolution layers, followed by temporal and spatial attention mechanisms, concluding with a residual addition. The temporal attention computes spatial averages of each channel and instant, followed by two 1 $\times$ 1 3D convolution layers and a Sigmoid activation. 
Similarly, the spatial attention layer employs a 3D convolution followed by Sigmoid activation. The RSSAB output is the addition of input and the spatial attention output. The final 3D convolution produces output data sized ($n_b \times 2 \times n_t \times n_{lat} \times n_{lon}$), where the 2 channels correspond to the wind's u and v components at a 10-metre height.

\section{Our Study}
\label{sec:case}
\subsection{Data Description}
\label{sec:case:datadescrip}
\subsubsection{DPIRD observation data source}

\begin{figure}[tbp]
    \centering
    \includegraphics[width=1.0\textwidth]{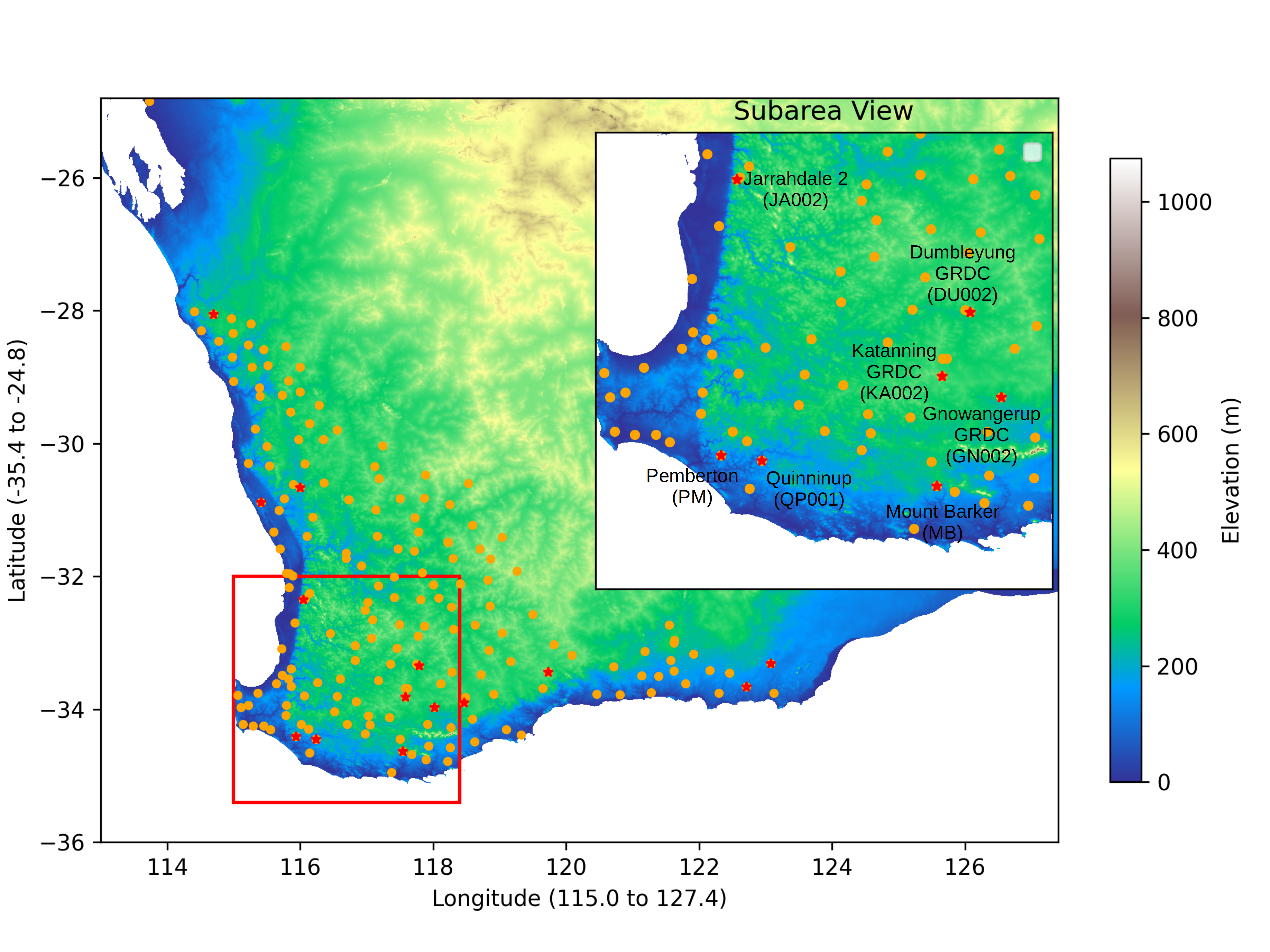} 
    \caption{Weather station locations in the southwest of Western Australia. The red square highlights the area used with 73 weather stations.}
    \label{fig:map}
\end{figure}

The observation data was retrieved using the publicly available Weather API 2.0 provided by the Western Australian Government's Department of Primary Industries and Regional Development (DPIRD)~\cite{dpridapi2024}. For this study, our focus was on the southwest area of Western Australia, as depicted in the delimited area (red square) shown in Figure \ref{fig:map}, which encompasses 73 weather stations marked by orange (3m stations) and red (10m stations) dots.

As discussed above, wind speed and direction at a 3-metre height, air temperature (T) and humidity (H), are pivotal variables for wind forecasting at a 10-metre height. Consequently, we gathered these four observations from 73 weather stations. To enhance the model's ability to learn the spatial correlation among these features, we transformed the wind speed and direction at a 3-metre height into their respective u and v components ($u_3$ and $v_3$).

Of the 73 weather stations, only seven stations provide wind speed and direction data at a height of 10 metres, highlighted by red circles in Figure \ref{fig:map}. The selected weather stations and their corresponding codes are as follows: Dumbleyung GRDC (DU002), Gnowangerup GRDC (GN002), Jarrahdale 2 (JA002), Katanning GRDC (KA002), Mount Barker (MB), Pemberton (PM), and Quinninup (QP001). Subsequently, the u and v components of the wind at a 10-metre height ($u_{10}$ and $v_{10}$) from these stations were utilised as the real labels for the model.

Data for the 4 features (T, H, $u_3$, $v_3$) from all weather stations\footnote{The 10-metre wind data of Nyabing GRDC (NY002) was unavailable for use due to inaccurate measurements since April 2022. The 3-metre wind direction data of Pemberton (PM) required manual correction, necessitating a 90-degree clockwise adjustment due to a fault in its wind probe.}, as well as the real labels ($u_{10}$ and $v_{10}$) for the 7 stations, were collected at 15-minute intervals, spanning from January 1, 2022, 00:00:00 UTC+0 to December 31, 2024, 23:45:00 UTC+0.

\subsubsection{ECMWF forecasting data source}
We also used the European Centre for Medium-Range Weather Forecasts (ECMWF) reanalysis dataset ERA5\footnote{ECMWF data was sourced from the ERA5 reanalysis dataset on single levels using the Climate Data Store (CDS) API~\cite{era5}.} as a part of the data source. The publicly available ERA5 dataset provides comprehensive information on various meteorological parameters, including wind components and mean sea level pressure, at a resolution of 0.25$^{\circ}$ $\times$ 0.25$^{\circ}$ (approximately 27.75 km $\times$ 27.75 km). 

We retrieved hourly data for the variables from January 1, 2022, 00:00:00 UTC+0 to December 31, 2024, 23:45:00 UTC+0, which are 10m\_u\_component\_of\_wind ($u_{10}\text{f}$): the eastward component of the wind velocity at a height of 10 metres above ground level; 10m\_v\_component\_of\_wind ($v_{10}\text{f}$): the northward component of the wind velocity at a height of 10 metres above ground level; mean\_sea\_level\_pressure (msl): the atmospheric pressure at mean sea level. 

These variables were chosen as they are fundamental for wind forecasting and play crucial roles in determining wind patterns and atmospheric conditions. The data was obtained on a target grid, as described in Section \ref{sec:exp:dataresolution}, ensuring consistency and compatibility with our modelling approach.

\subsubsection{Terrain data source}
In addition to the DPIRD observation data and the ECMWF forecasting data, we incorporated terrain data into our modelling framework to account for the influence of topography on wind patterns, as shown in Fig \ref{fig:map}. The terrain data used in this study is the Digital Elevation Model (DEM) Version 3~\cite{geoscienceaustralia2008}, which is publicly available and covers the whole of Australia. The DEM dataset provides a grid of ground-level elevation points with a grid spacing of 9 seconds in latitude and longitude, corresponding to approximately 0.25 kilometres in the GDA94 coordinate system. By integrating terrain data into our modelling approach, we aim to enhance the accuracy and reliability of our wind forecasts by accounting for the complex interactions between terrain features and atmospheric processes.

\subsubsection{Add time features}
Apart from the eight features obtained from the DPIRD weather stations, ECMWF forecasting and the terrain, another six time-related features were introduced to enhance the models' understanding of temporal patterns and seasonality. These features included the sine and cosine components of the month within a year, hour within a day, and day within a year.

\subsection{Experiments configuration}
\subsubsection{Data resolution}
\label{sec:exp:dataresolution}

The selected subarea was gridded ranging from latitude -32$^\circ$ to -35.4$^\circ$ and longitude 115$^\circ$ to 118.4$^\circ$ (approximately 140,000 $\text{km}^2$), with a cell size of 0.1$^\circ$ in both latitude and longitude (121 $\text{km}^2$). This subarea was chosen strategically as it encompasses seven weather stations (JA002, DU002, KA002, GN002, PM, QP001, and MB) with labelled wind components ($u_{10}$ and $v_{10}$ observations), and 73 weather stations with $u_3$ and $v_3$ observations, facilitating comprehensive performance evaluation.

All features and labels were binned into the grid using specific methods tailored to their data type. For DPIRD sourced data, including temperature (T), humidity (H), and wind components at 3 metres height ($u_3$ and $v_3$), a nearest neighbour approach was employed due to their sparse distribution within the subarea. Empty cells were subsequently assigned zeroes for these four features to maintain consistency in the dataset. ECMWF forecasting data underwent different processing, being hourly and distributed on a grid with a 0.25$^\circ$ resolution. We applied linear interpolation along the spatial dimension to align with the model's temporal granularity and enhanced spatial resolution, and resampled the data for every 15-minute interval. The terrain data was characterised by its exceptionally high resolution of 9 seconds. The nearest terrain data point, positioned closest to the cell's centre, was selected for each cell within the grid.

\subsubsection{Training and testing data samples}
We devised a specific temporal arrangement for the model training and testing to ensure robust evaluation and validation. The dataset's temporal scope spans from January 1, 2022, 00:00:00 UTC+0 to December 31, 2023, 23:45:00 UTC+0, with an interval of 15 minutes. 
To establish a balanced evaluation protocol, the data for the final five days of each month were allocated to the testing set, while the remaining data was designated for model training. 

\begin{figure}[tbp]
    \centering
    \includegraphics[width=1.0\textwidth]{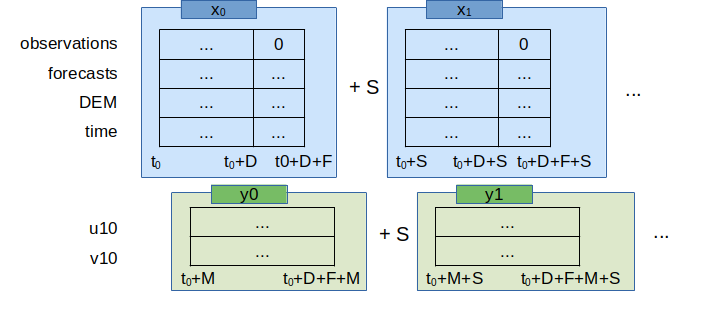} 
    \caption{Input and output data samples}
    \label{fig:sample}
\end{figure}

Figure \ref{fig:sample} illustrates the temporal structure of the input ($x_n$) and output ($y_n$) data sequences.
Here, $t_0$ denotes the starting time of the input sample, with D representing the duration (2 days) of each sample. F signifies the forward window (4 hours), while M denotes the moving window of $y$ (4 hours), and S indicates the sliding window for the next sample (15 minutes).

The input data ($x$) encompasses 14 features, categorised into four classes: observations (T, H, $u_3$, $v_3$), forecasts ($u_{10}\text{f}$, $v_{10}\text{f}$, msl), terrain (DEM), and six time-related features. The output data ($y$) consists of $u_{10}$ and $v_{10}$ components. 
We configure the model to predict $u_{10}$ and $v_{10}$ starting from the time $t_0$+D; hence, any observations occurring after $t_0$+D are set to zero in the input sample.

To optimise the utilisation of ECMWF forecast data, we extend the forward window by F hours. Consequently, the end time of the input sample becomes $t_0$+D+F. The output $y$ retains the same length as $x$, with the start time shifted M units ahead of $x$. This process ensures alignment between the input and output sequences.

Subsequent pairs of $x$ and $y$ are generated iteratively, following the same procedure of advancing the start time of S relative to $t_0$. This systematic approach ensures the continuity and coherence of the input-output data pairs throughout the temporal sequence.

\subsubsection{Loss function}

The loss function is calculated as the mean squared error (MSE) between the predicted $u_{10}$ and $v_{10}$ values ($\widehat{u_{10}}$ and $\widehat{v_{10}}$) and the actual label values for $u_{10}$ and $v_{10}$ at the 7 labelled stations (\ref{eq:loss1}).

\begin{equation}
\begin{split}
\label{eq:loss1}
L = \frac{1}{N} \sum_{i=1}^{N} \left( \frac{1}{L} \sum_{j=1}^{L} \right. & \left. \left( \frac{1}{2} \left( u_{10_{ij}}-\widehat{u_{10_{ij}}} \right)^2 \right. \right. \\
& \left. \left. + \frac{1}{2} \left(v_{10_{ij}}-\widehat{v_{10_{ij}}} \right)^2 \right) \right)
\end{split}
\end{equation}

where N is the total number of weather stations with real labels within the area, which is seven in the current case; L is the number of data points in a sample which is D + F; $u_{10_{ij}}$ and $v_{10_{ij}}$ represents the true labels at station $i$ and time $j$; $\widehat{u_{10_{ij}}}$ and  $\widehat{v_{10_{ij}}}$ denotes the predicted labels at station $i$ and time $j$.

\subsection{Performance metrics}
We employ two evaluation approaches to assess the performance of our wind forecasting model, given the availability of real labels ($u_{10}$, $v_{10}$) from the seven stations within the subarea of interest. All evaluations are conducted considering the prediction horizon of eight hours.

\subsubsection{Evaluation on inclusive labelled stations}
In this approach, we evaluate Mean Absolute Error (MAE) and Root Mean Square Error (RMSE) (see Eq. \eqref{mae} and Eq. \eqref{rmse}) between our predictions ($\widehat{u_{10}}$, $\widehat{v_{10}}$) and the real labels, as well as between the ECMWF forecasting and the real labels, focusing on the seven stations within the subarea. These metrics are computed for the u and v components, speed, sine and cosine for the wind, respectively. The evaluation is performed across the entire 2-year testing dataset and is further stratified for the winter (June to September) and summer (November to February) seasons, as well as for daytime (5 am to 7:30 pm UTC+8 in summer and 7:15 am to 5:30 pm UTC+8 in winter) and nighttime periods. 

\begin{equation}
\label{mae}
MAE = \frac{1}{n} \sum_{i=1}^{n} |y_i - \hat{y}_i|
\end{equation}

\begin{equation}
\label{rmse}
RMSE = \sqrt{\frac{1}{n} \sum_{i=1}^{n} (y_i - \hat{y}_i)^2}
\end{equation}

where $n$ is the number of observations, $y_i$ is the actual value ($u_{10}$, $v_{10}$), and $\hat{y}_i$ is the predicted value ($\widehat{u_{10}}$, $\widehat{v_{10}}$).

\subsubsection{Evaluation on unlabelled stations}

In the 140,000 $\text{km}^2$ subarea under consideration, the availability of only 7 labelled stations with observation data for $u_{10}$ and $v_{10}$, sparsely distributed across the region poses a significant challenge for model evaluation on any other coordinates. However, our observations reveal a strong correlation between $u_{10}$, $v_{10}$ and their counterparts at 3 metres ($u_3$ and $v_3$), as evidenced in Table \ref{tab:coruv_10_3} for the 7 labelled stations. Leveraging this correlation, there arises an opportunity to assess model performance at the unlabelled stations by examining the correlation coefficients between predicted $\widehat{u_{10}}$ and observed $u_3$, as well as between predicted $\widehat{v_{10}}$ and observed $v_3$, as delineated in Eq. \eqref{eq:r_uv10_uv_3}, where $y_{10}$ and $y_3$ represents the $u$ and $v$ components of the wind at 10 metres and 3 metres height respectively, $\text{cov}$ denotes the covariance and $\text{var}$ represents of variance. This approach holds promise for extending model performance evaluation across the broader region with greater confidence.

\begin{table}
\centering
\caption{\textbf{Correlations between 10-metre and 3-metre wind components} \label{tab:coruv_10_3}}
\resizebox{\linewidth}{!}{%
\begin{tblr}{
  vline{2} = {-}{},
  hline{2} = {-}{},
}
site & DU002 & GN002 & JA002 & KA002 & MB   & PM   & QP001 \\
$r(u_{10},u_3)$ & 0.99  & 0.98  & 0.88  & 0.98  & 0.98 & 0.98 & 0.99  \\
$r(v_{10},v_3)$ & 0.99  & 0.95  & 0.98  & 0.98  & 0.98 & 0.97 & 0.97  
\end{tblr}
}
\end{table}

\begin{equation}
\label{eq:r_uv10_uv_3}
r(\widehat{y_{10}}, y_3) = \frac{\text{cov}(\widehat{y_{10}}, y_3)}{\sqrt{\text{var}(\widehat{y_{10}}) \cdot \text{var}(y_3)}}
\end{equation}

\section{Results and discussion}
\label{sec:result}
\subsection{Model performance on labelled stations}
\begin{figure}[tbp]
    \centering
    \includegraphics[width=0.8\textwidth]{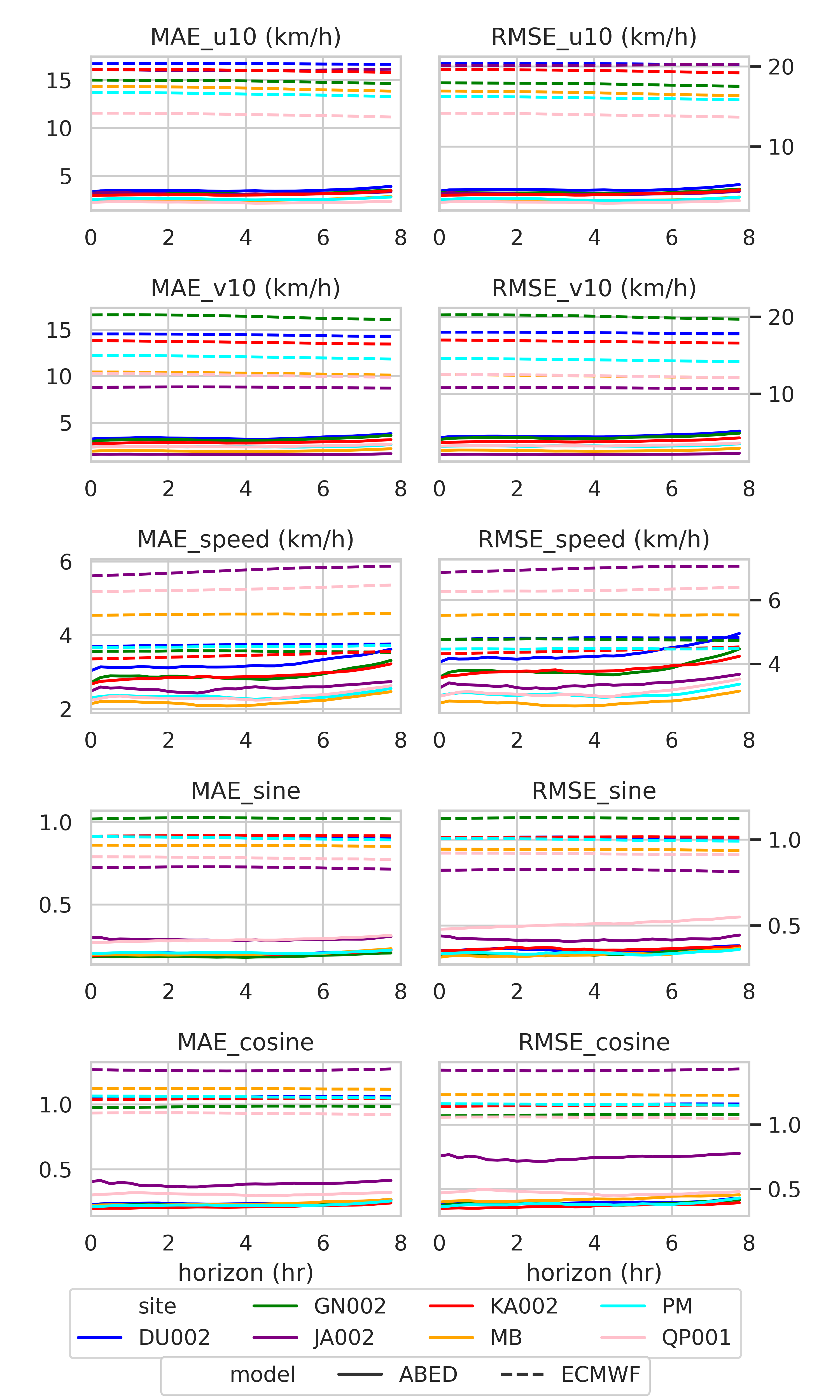} 
    \caption{MAE and RMSE of $u_{10}$ and $v_{10}$ at the 7 labelled stations from Jan-2022 to Dec-2023}
    \label{fig:7sites_all}
\end{figure}

\begin{figure}[tbp]
    \centering
    \includegraphics[width=0.8\textwidth]{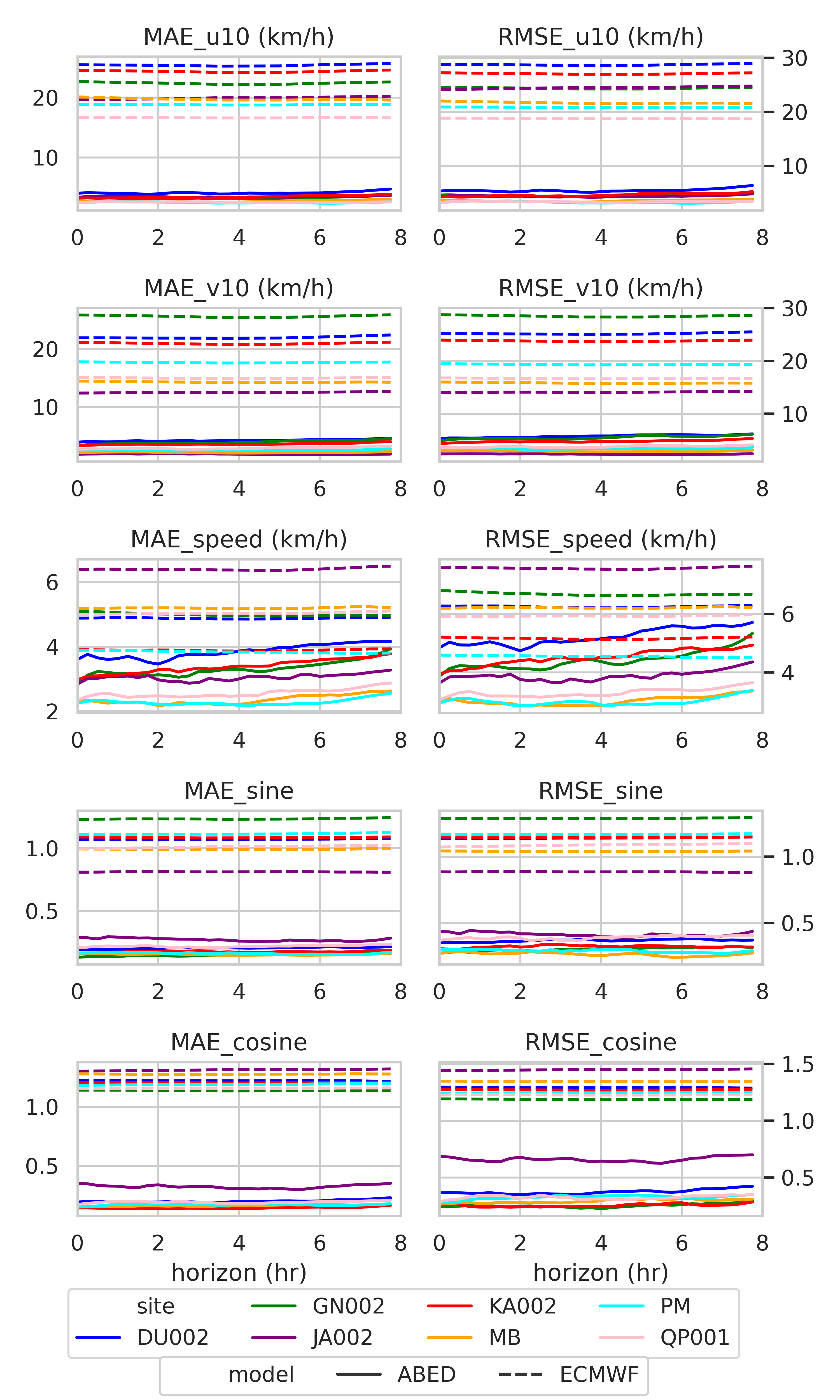} 
    \caption{MAE and RMSE of $u_{10}$ and $v_{10}$ at the 7 labelled stations during the period of summer daytime from Jan-2022 to Dec-2023}
    \label{fig:7sites_summerday}
\end{figure}

\begin{figure}[tbp]
    \centering
    \includegraphics[width=0.8\textwidth]{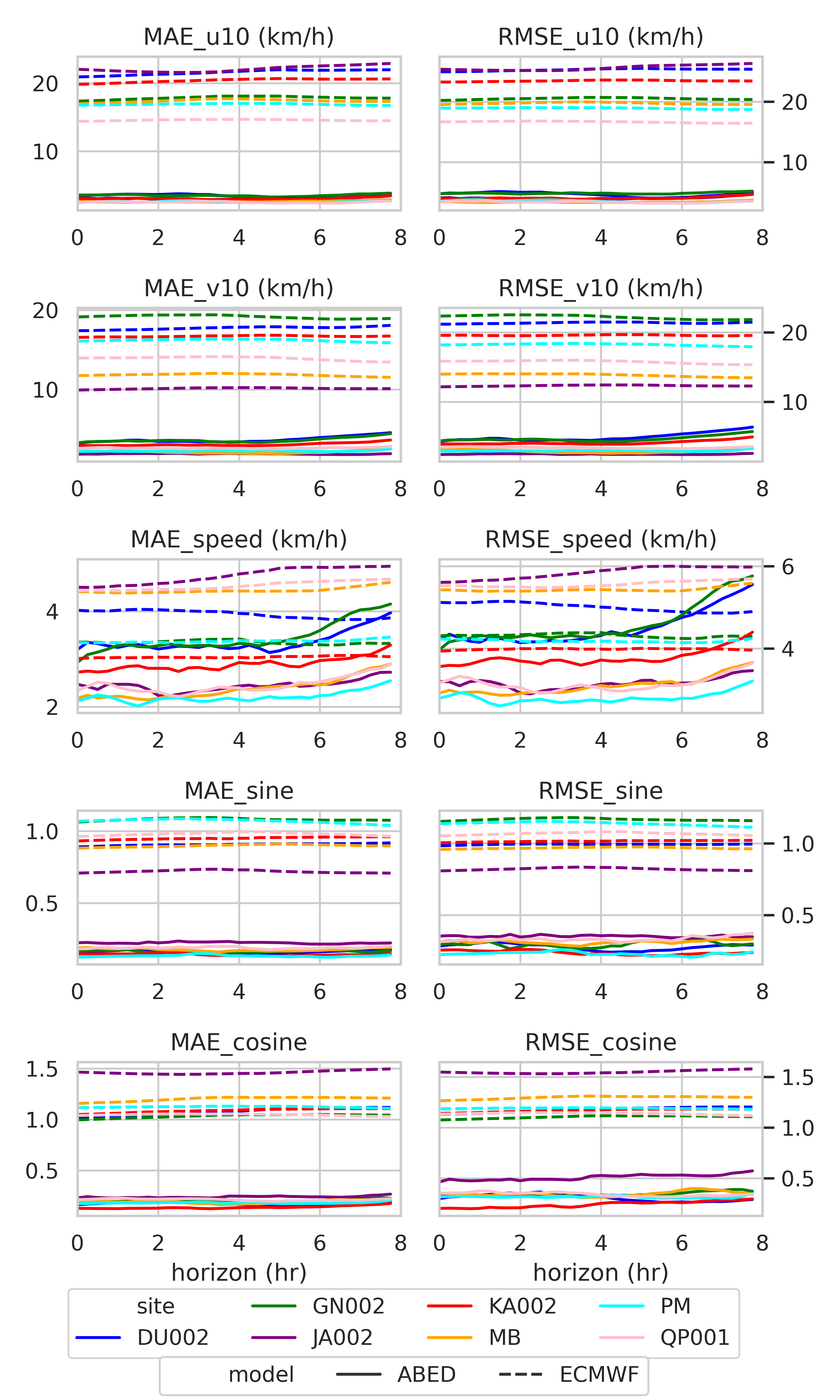} 
    \caption{MAE and RMSE of $u_{10}$ and $v_{10}$ at the 7 labelled stations during the period of summer nighttime from Jan-2022 to Dec-2023}
    \label{fig:7sites_summernight}
\end{figure}

\begin{figure}[tbp]
    \centering
    \includegraphics[width=0.8\textwidth]{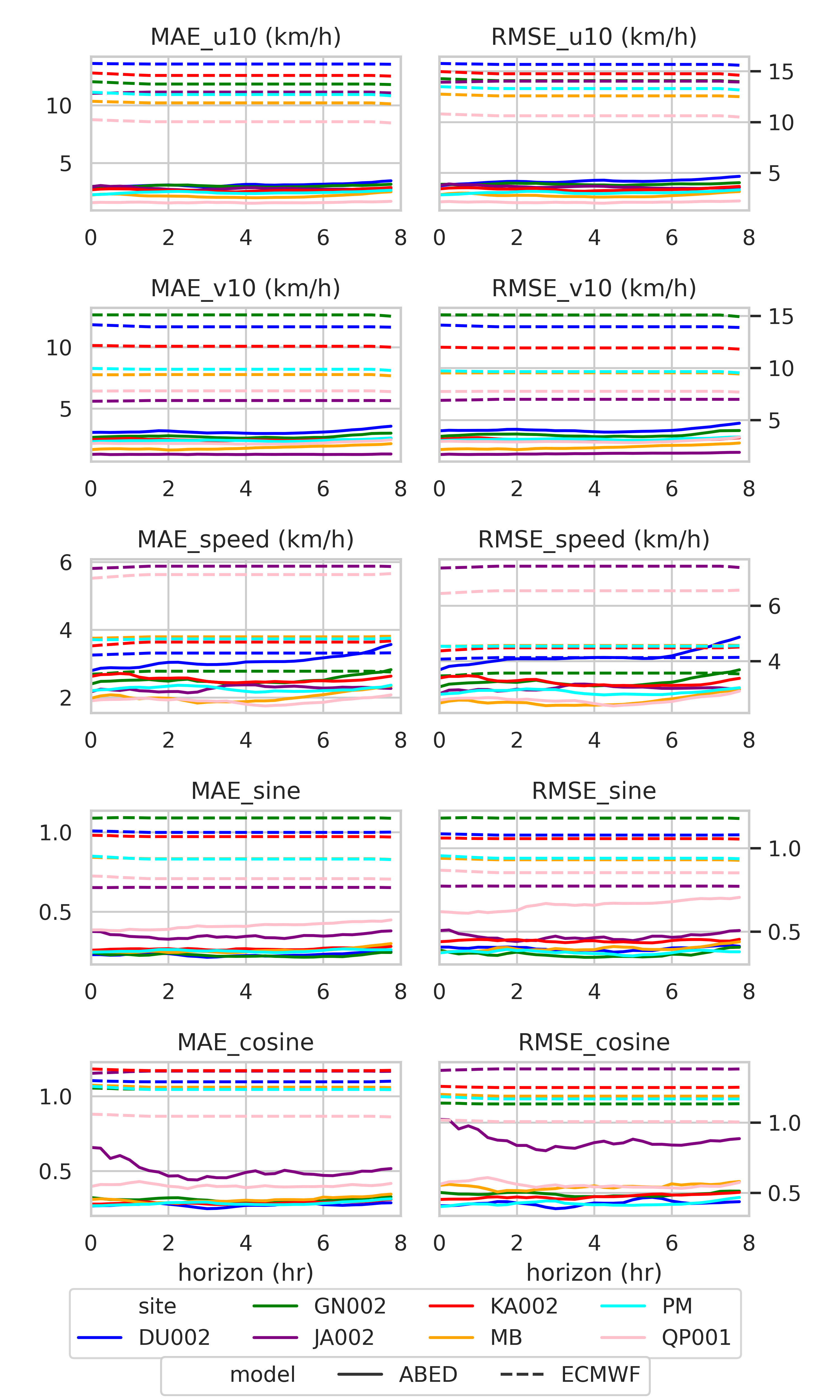} 
    \caption{MAE and RMSE of $u_{10}$ and $v_{10}$ at the 7 labelled stations during the period of winter daytime from Jan-2022 to Dec-2023}
    \label{fig:7sites_winterday}
\end{figure}

\begin{figure}[tbp]
    \centering
    \includegraphics[width=0.8\textwidth]{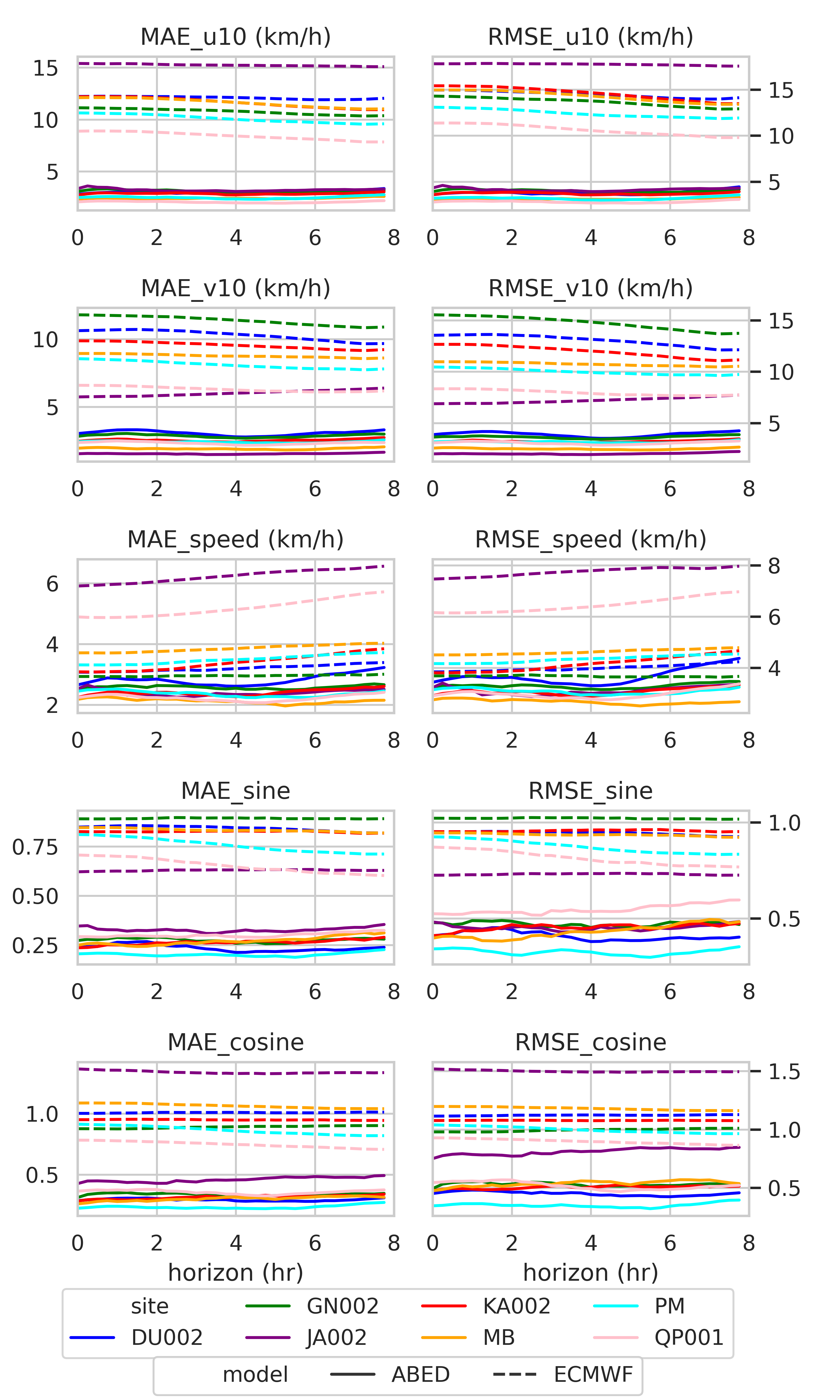} 
    \caption{MAE and RMSE of $u_{10}$ and $v_{10}$ at the 7 labelled stations during the period of winter nighttime from Jan-2022 to Dec-2023}
    \label{fig:7sites_winternight}
\end{figure}

The MAE and RMSE of the u and v components of the wind ($u_{10}$ and $v_{10}$) over the entire 2 years testing set are demonstrated in Fig \ref{fig:7sites_all} for the 7 inclusive labelled stations. More specifically, Fig \ref{fig:7sites_summerday}, \ref{fig:7sites_summernight}, \ref{fig:7sites_winterday} and \ref{fig:7sites_winternight} presented the performance for summer daytime, summer nighttime, winter daytime and winter nighttime. 

The proposed model predicts the $u_{10}$ and $v_{10}$ wind components better than ECMWF reanalysis across all 7 inclusive stations. It is important to note that the ECMWF reanalysis data includes assimilated observations at a single time stamp, rather than providing forecasts across multiple time horizons. Over the two-year testing set, the proposed model exhibits approximately a 200\% improvement in MAE and a 300\% improvement in RMSE for the u and v components. Specifically, using the proposed model, the MAE and RMSE for the u and v components remain consistently below 4.2 km/h and 5.7 km/h, respectively, across prediction horizons ranging from 15 minutes to 8 hours. In contrast, the MAE and RMSE values for ECMWF forecasts range from 9 to 17 km/h and 10 to 20 km/h, respectively. It is worth noting that the MAE and RMSE of the ECMWF reanalysis for the absolute value of wind speed are lower than 6 km/h and 8 km/h, respectively. However, the ECMWF wind direction performed poorly, which adversely affected the MAE and RMSE of the u and v components.

Comparison of model performance between summer daytime, nighttime, winter daytime, and nighttime reveals nuanced trends. The model consistently outperforms in winter conditions, with MAE values for the u and v components consistently below 4.16 km/h and RMSE values below 5.34 km/h. Conversely, predictions are slightly less accurate during summer periods. Interestingly, the model's performance remains consistent between day and night.

As expected, model performance diminishes with increasing prediction horizons. For very-short term predictions (15 minutes to 1 hour), the model achieves MAE values ranging from 1.38 to 3.1 km/h and RMSE values ranging from 1.86 to 4 km/h for the u and v components, respectively.

The proposed model predictions vary across different locations, with coastal stations such as QP001, PM, JA002, and MB exhibiting more favourable outcomes than inner stations like DU002, GN002, and KA002. In contrast, ECMWF reanalysis data shows greater performance for the inner stations compared to the coastal stations. This spatial variability underscores the importance of considering local geography and environmental factors when assessing model performance.

\begin{figure}[tbp]
    \centering
    \includegraphics[width=0.8\textwidth]{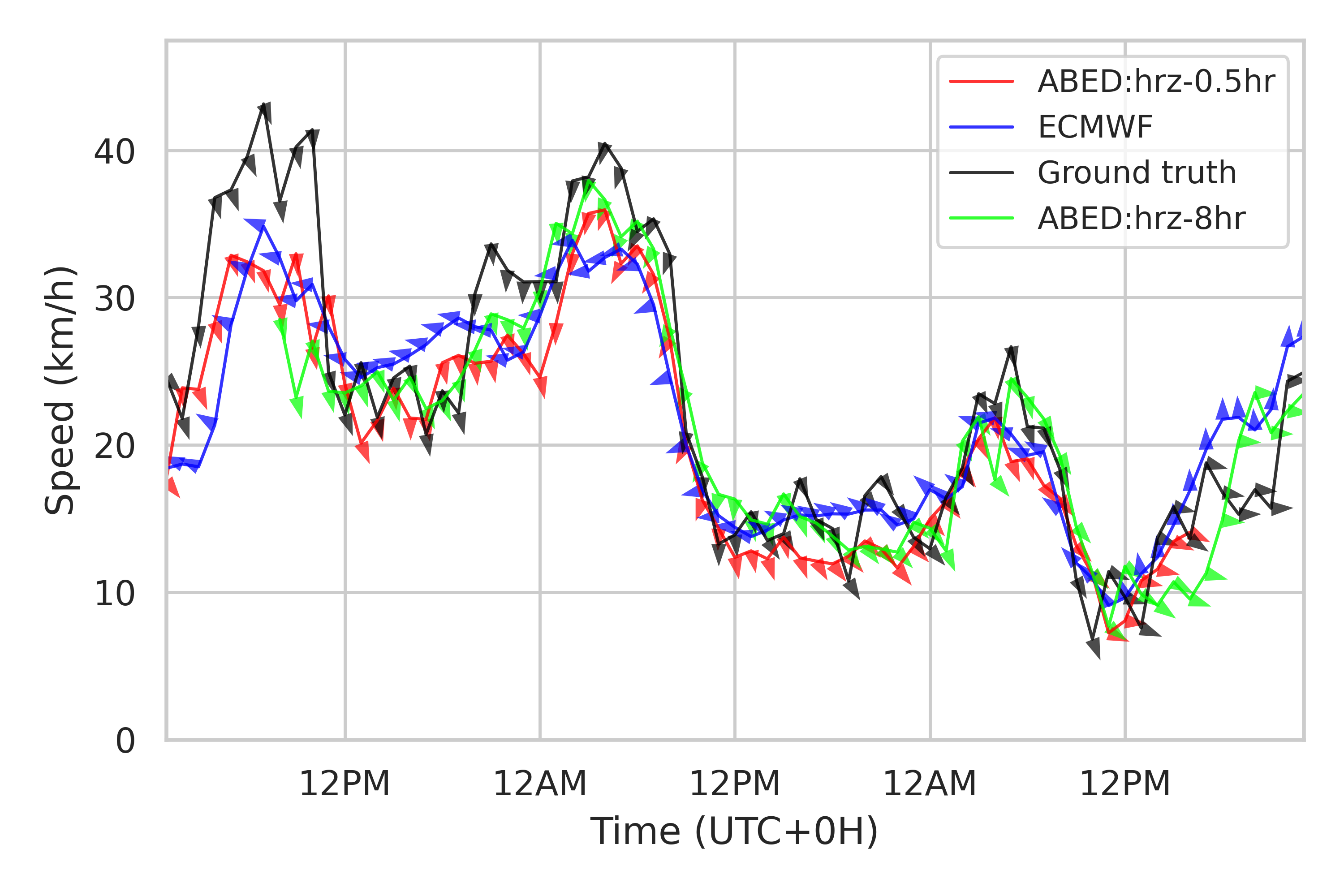} 
    \caption{GN002 - July 29 to July 31, 2022 (winter)}
    \label{fig:gnwinter}
\end{figure}

\begin{figure}[tbp]
    \centering
    \includegraphics[width=0.8\textwidth]{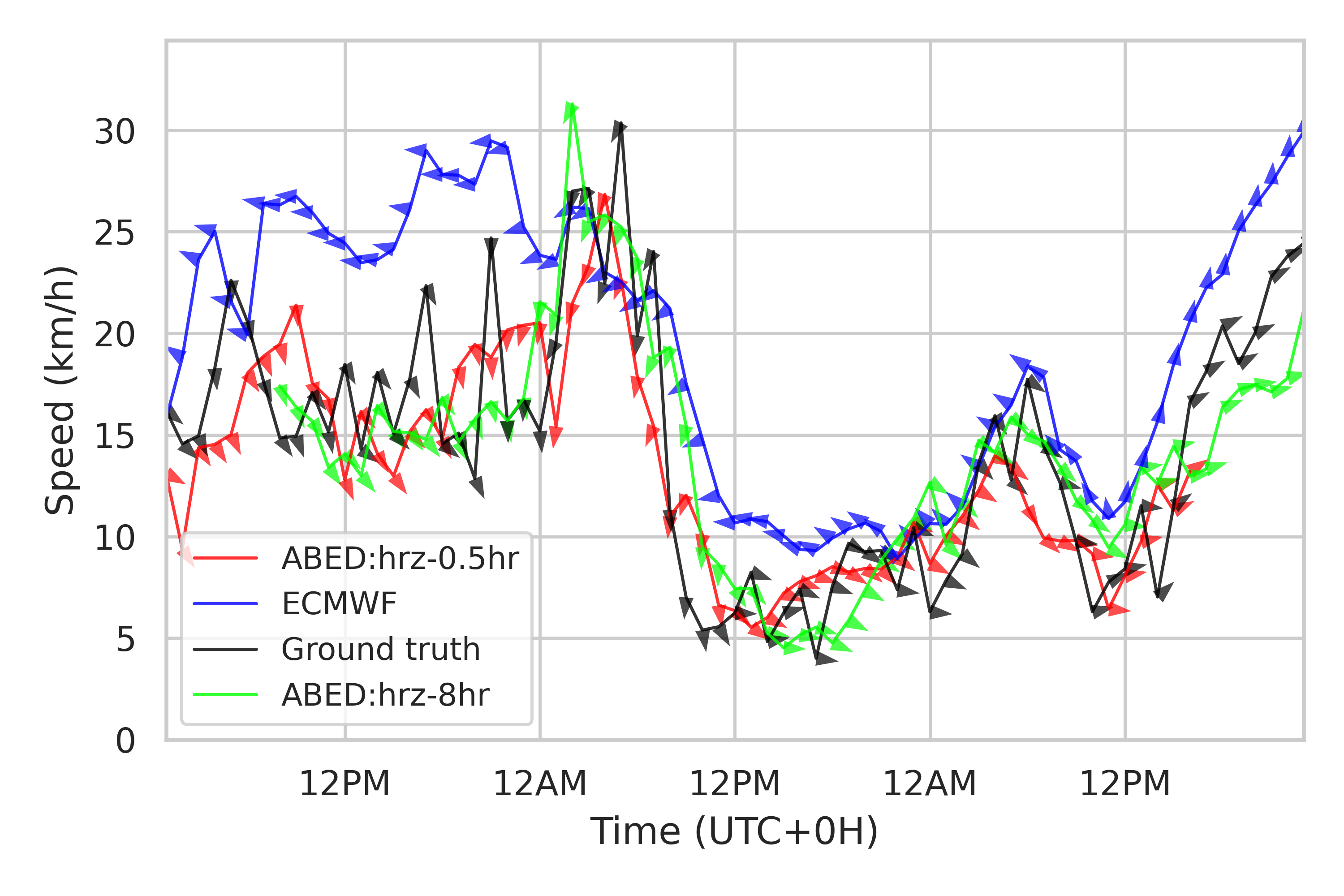} 
    \caption{PM - July 29 to July 31, 2022 (winter)}
    \label{fig:pmwinter}
\end{figure}

\begin{figure}[tbp]
    \centering
    \includegraphics[width=0.8\textwidth]{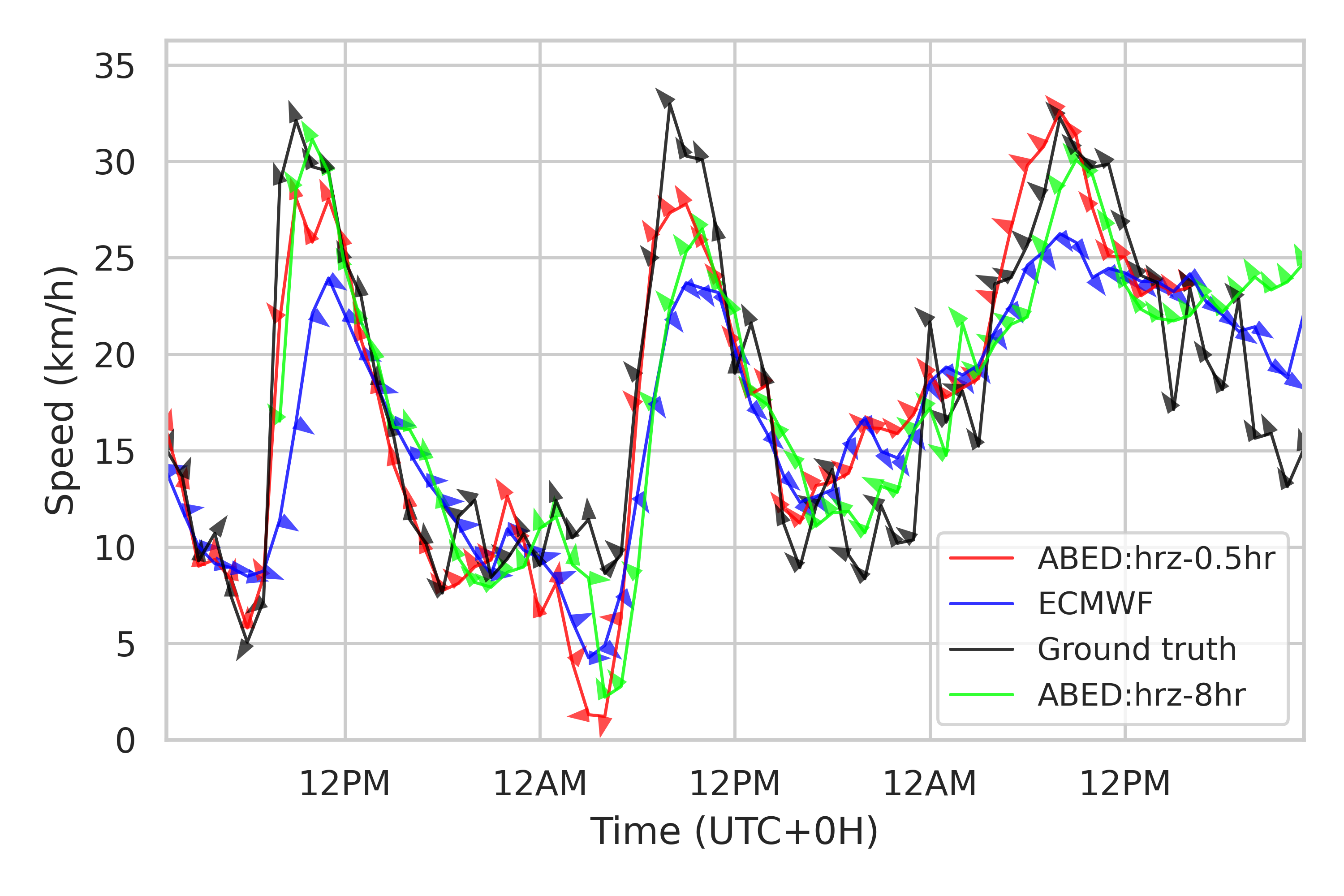} 
    \caption{GN002 - Jan 29 to Jan 31, 2023 (summer)}
    \label{fig:gnsummer}
\end{figure}

\begin{figure}[tbp]
    \centering
    \includegraphics[width=0.8\textwidth]{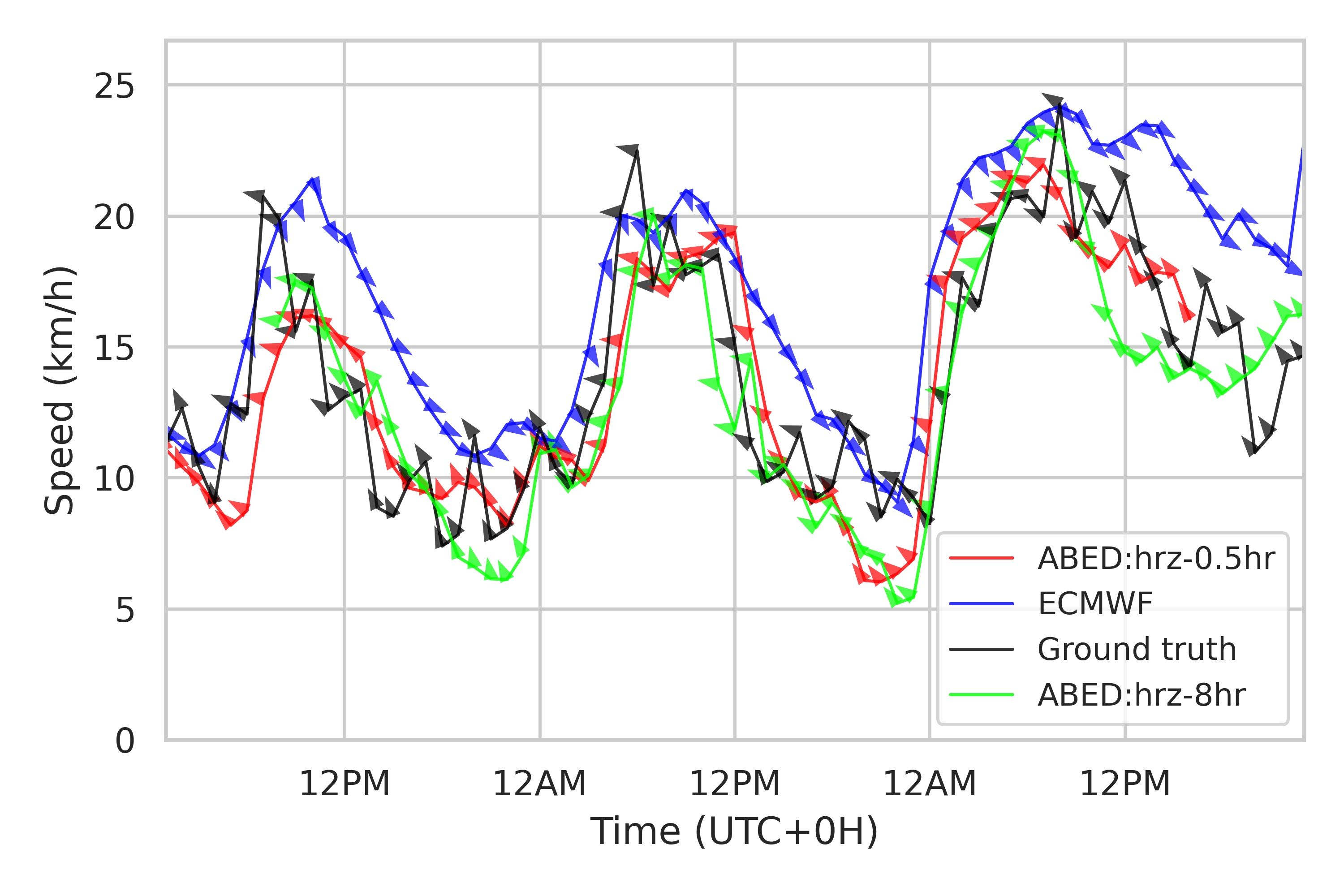} 
    \caption{PM - Jan 29 to Jan 31, 2023 (summer)}
    \label{fig:pmsummer}
\end{figure}

To further scrutinise the model's performance across different sites, horizons, and seasons, Fig \ref{fig:gnwinter}, \ref{fig:pmwinter}, \ref{fig:gnsummer}, and \ref{fig:pmsummer} depict wind speed and direction plots for two representative stations: GN002 (an inland station) and PM (a coastal station). The plots span two distinct time periods: from July 28 to July 31, 2022 (winter) and from January 28 to January 31, 2023 (summer). 
Two prediction horizons are considered: 30 minutes (short-term) and 8 hours (mid-term). 
The lines represent the absolute wind speed calculated as $\sqrt{u_{10}^2+v_{10}^2}$ in km/h, and the arrows indicate the wind direction. 

The ECMWF reanalysis exhibit an underestimation problem for GN002 during the summer months. Additionally, the ECMWF struggles to accurately capture wind direction. These limitations could have detrimental implications for various applications, such as wind energy planning, agriculture, and transportation safety.

\subsection{Model performance on unlabelled stations}
\begin{figure}[tbp]
    \centering
    \includegraphics[width=0.6\textwidth]{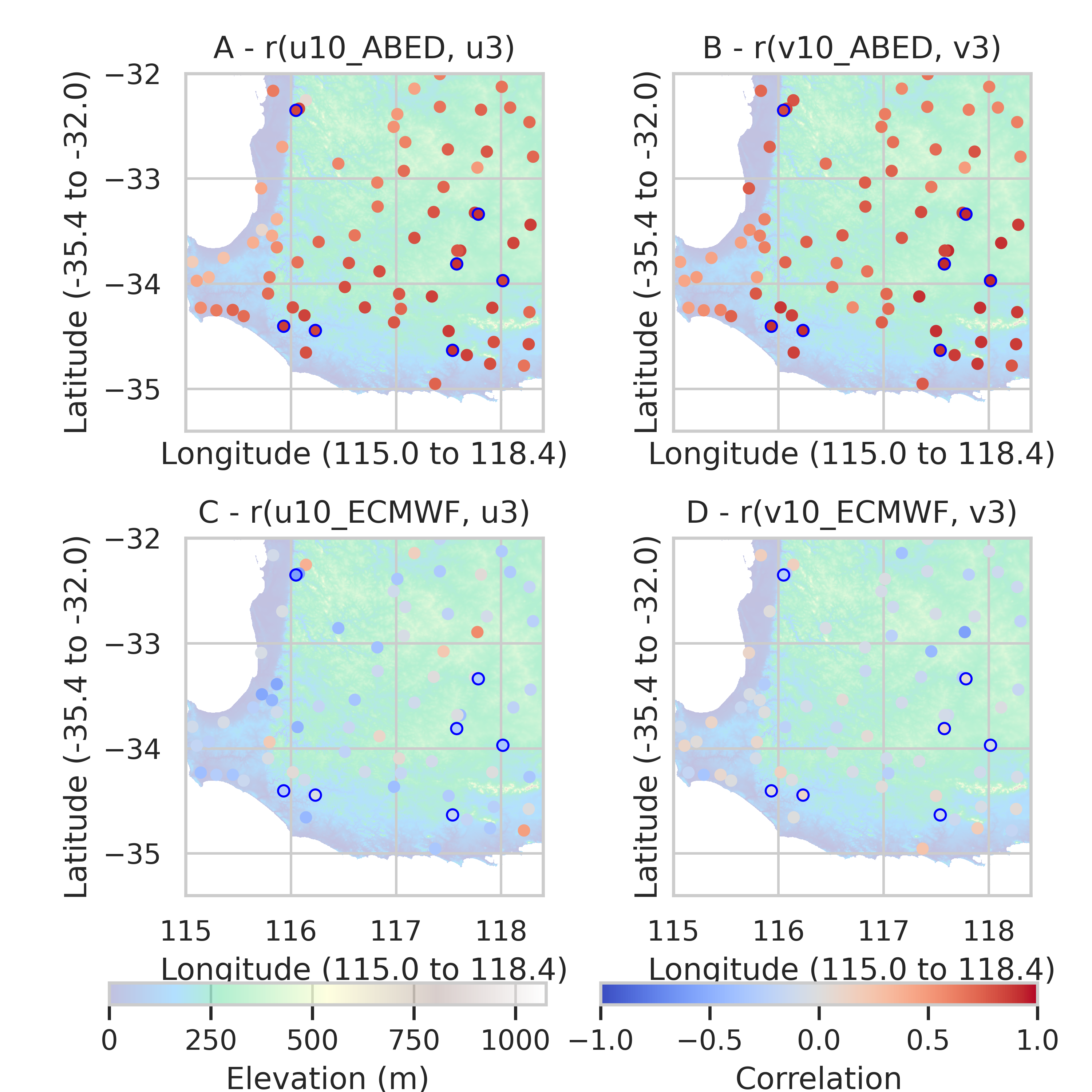} 
    \caption{Correlations between ${uv_{10}}$ and $uv_{3}$ - horizon = 0.5 hour}
    \label{fig:cor0.5}
\end{figure}

\begin{figure}[tbp]
    \centering
    \includegraphics[width=0.6\textwidth]{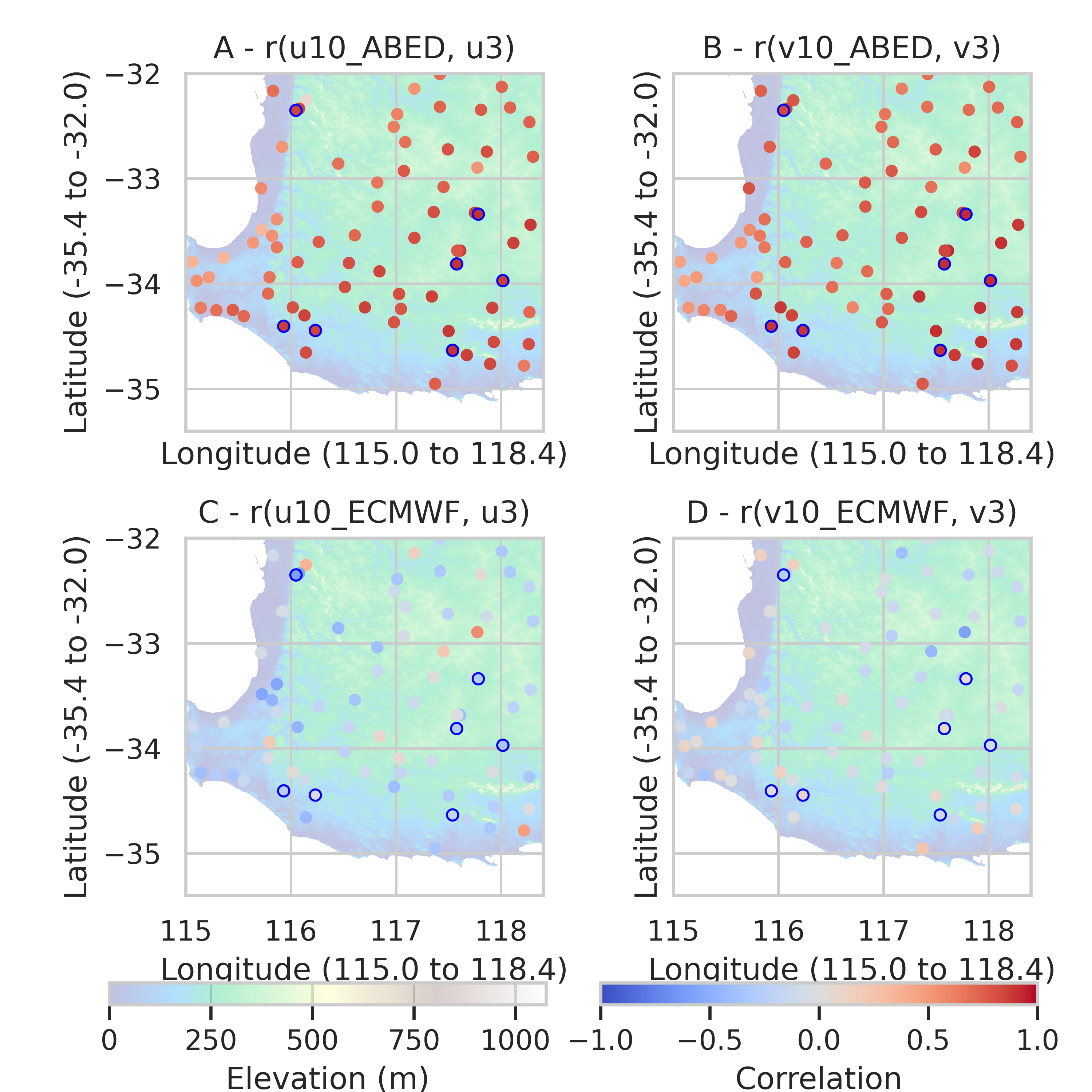} 
    \caption{Correlations between predicted ${uv_{10}}$ and $uv_{3}$ - horizon = 2 hours}
    \label{fig:cor2}
\end{figure}

\begin{figure}[tbp]
    \centering
    \includegraphics[width=0.6\textwidth]{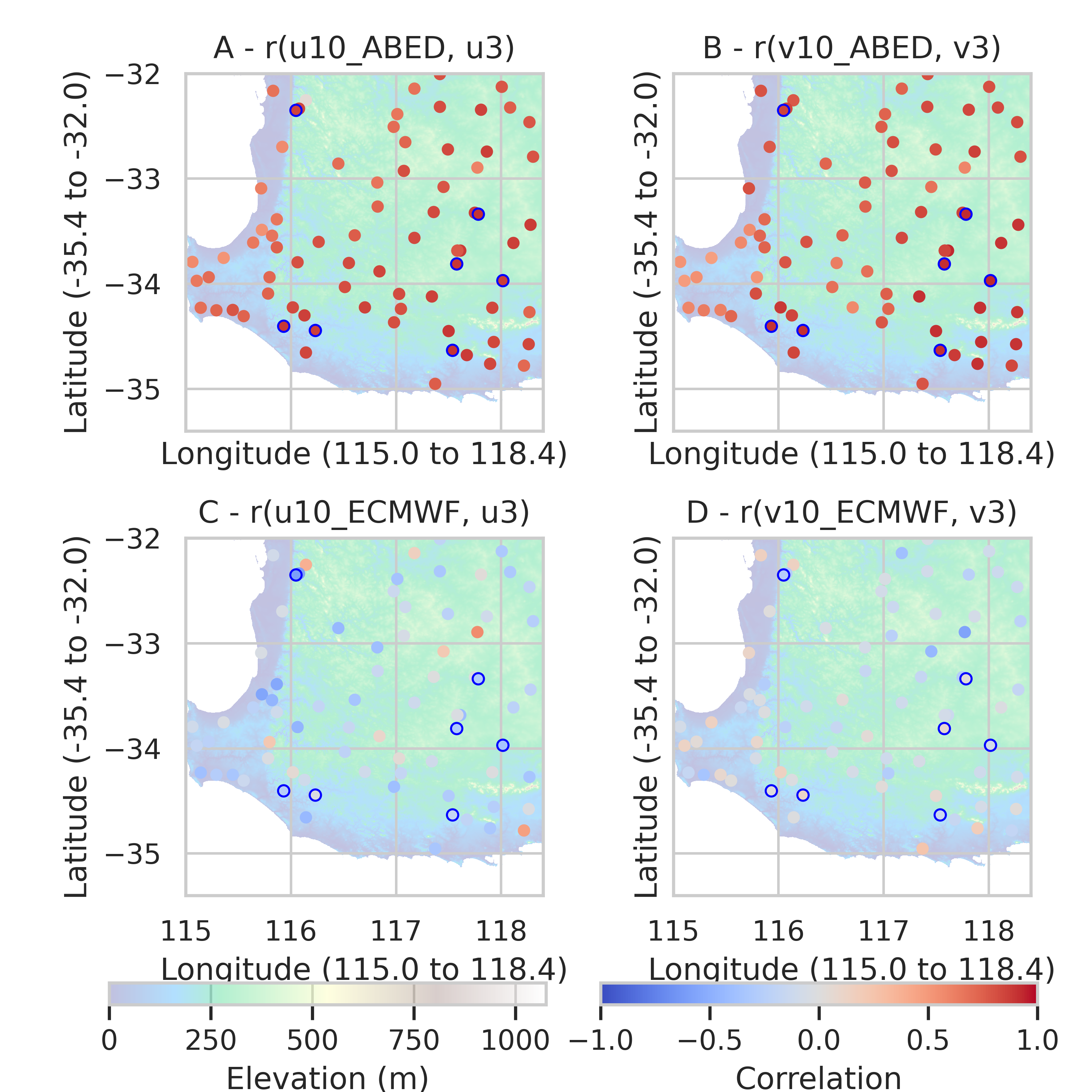} 
    \caption{Correlations between predicted ${uv_{10}}$ and $uv_{3}$ - horizon = 4 hours}
    \label{fig:cor4}
\end{figure}

\begin{figure}[tbp]
    \centering
    \includegraphics[width=0.6\textwidth]{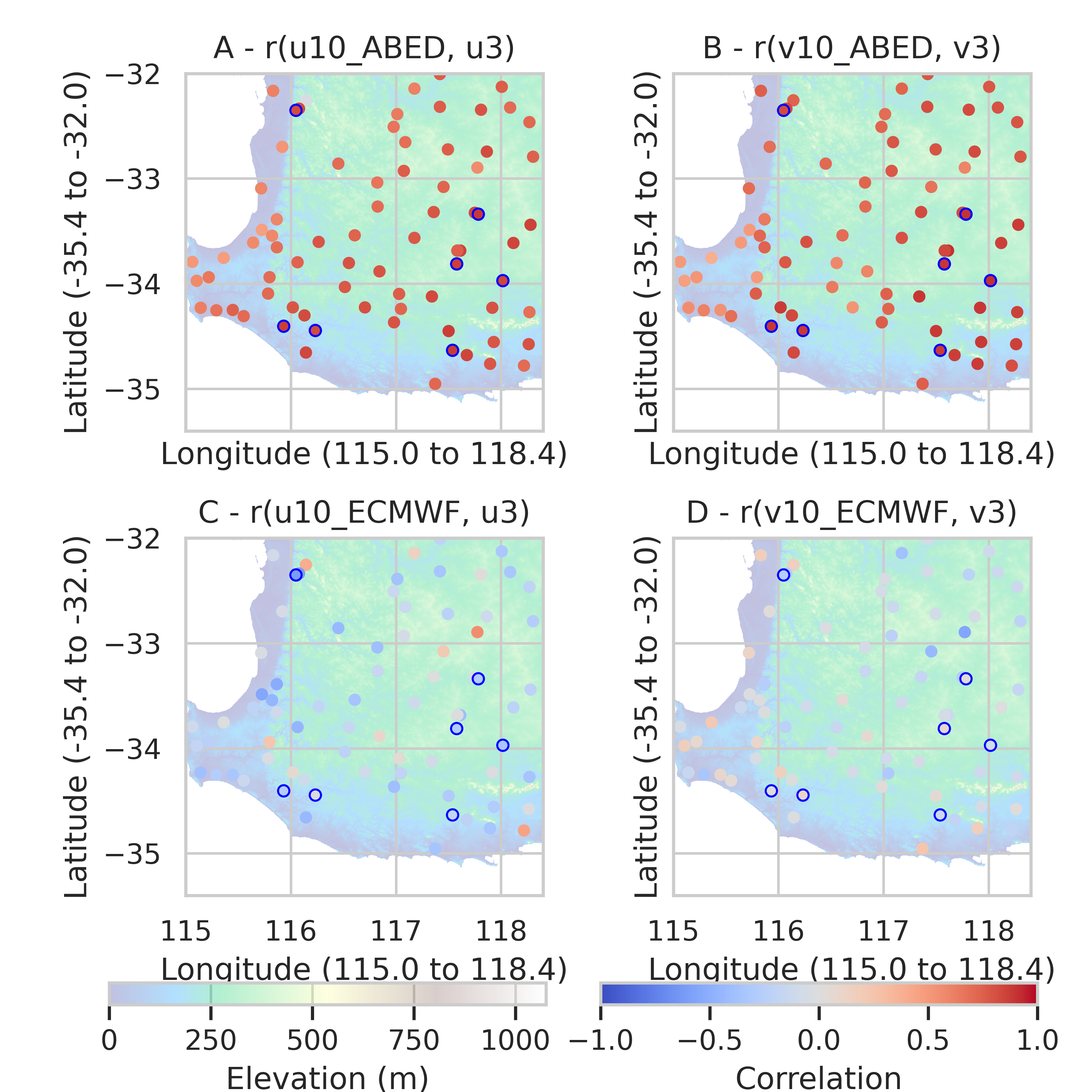} 
    \caption{Correlations between predicted ${uv_{10}}$ and $uv_{3}$ - horizon = 8 hours}
    \label{fig:cor8}
\end{figure}

Fig \ref{fig:cor0.5}, \ref{fig:cor2}, \ref{fig:cor4}, and \ref{fig:cor8} illustrate the correlations between the forecasts of $u_{10}$ and $v_{10}$ and the observations of $u_3$ and $v_3$ across the year of 2022 and 2023, considering horizons of 0.5 hour, 2 hours, 4 hours, and 8 hours. The top rows (``A'' and ``B'') represent the predictions generated by the proposed model for $\widehat{u_{10}}$ and $\widehat{v_{10}}$, while the bottom rows (``C'' and ``D'') depict the ECMWF forecasts for the same variables. The blue rings denote the 7 labelled stations.

The comparative analysis clearly demonstrates the proposed model's superior performance over ECMWF forecasts, evidenced not only at the 7 labelled stations but also across the broader region. The proposed model exhibits robust correlations with the 3-metre wind profile, which underscores the high reliability and trustworthiness of the proposed model's forecast data throughout the entire geographical area.

\subsection{Forecast on the area}
\begin{figure}[tbp]
    \centering
    \includegraphics[width=0.8\textwidth]{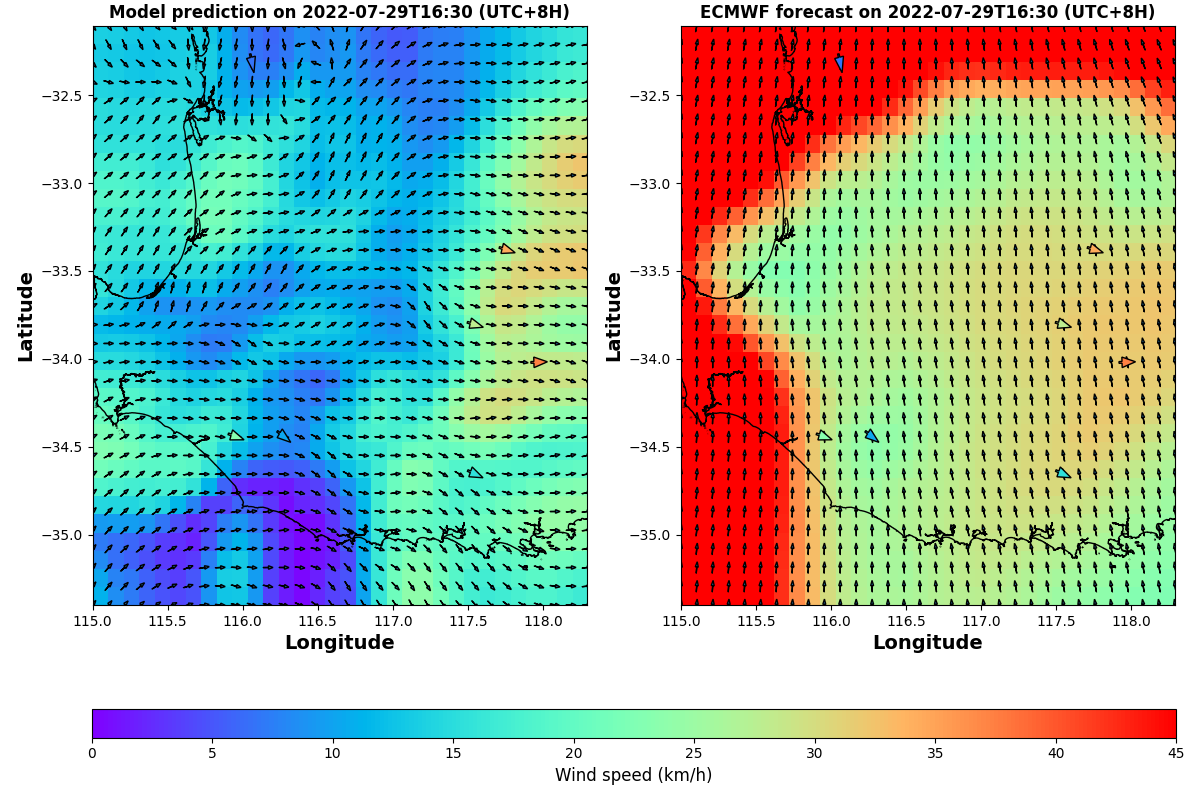} 
    \caption{Wind forecast on 2022-07-29 16:30 (UTC+8)}
    \label{fig:areaforecast_winter}
\end{figure}

\begin{figure}[tbp]
    \centering
    \includegraphics[width=0.8\textwidth]{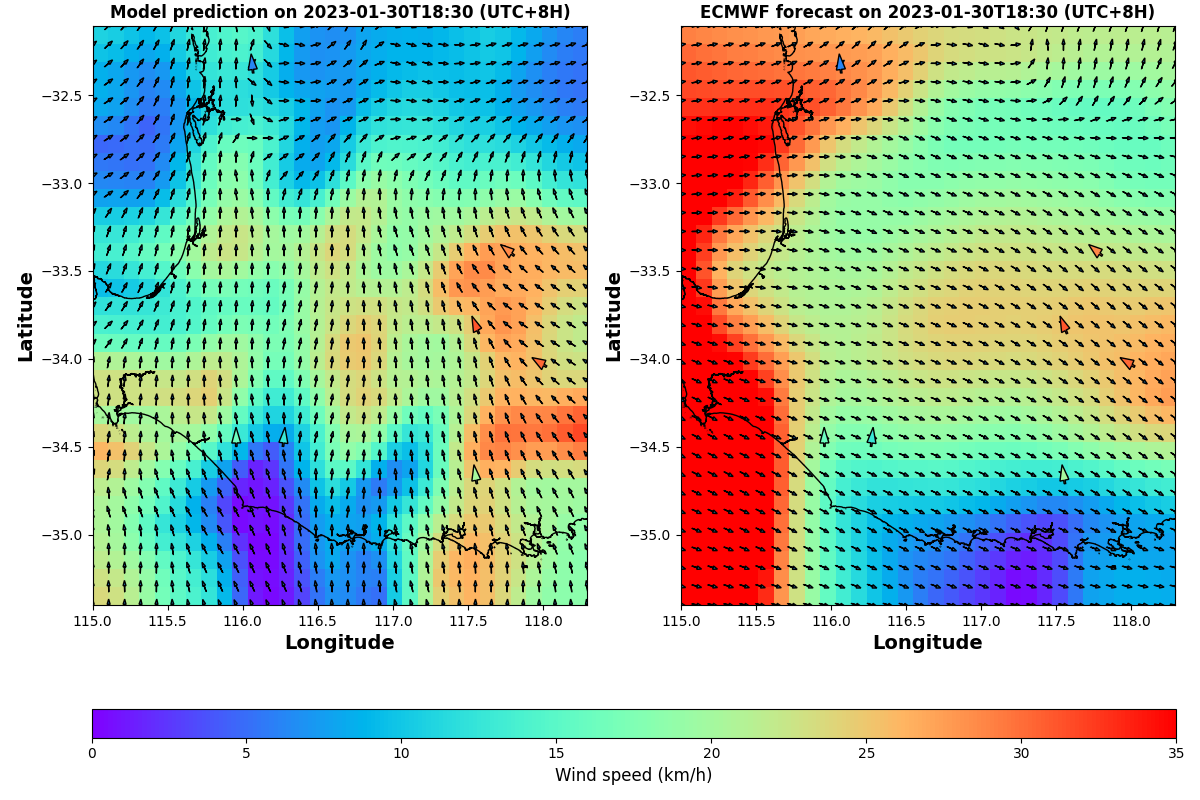} 
    \caption{Wind forecast on 2023-01-30 18:30 (UTC+8)}
    \label{fig:areaforecast_summer}
\end{figure}

Fig \ref{fig:areaforecast_winter} and \ref{fig:areaforecast_summer} depict the wind forecast for the selected area on a winter day and a summer day at a horizon of 30 minutes. In these figures, the left plot illustrates the forecasts generated by the proposed model, while the right plot presents the forecasts from the ECMWF model. The enlarged arrows represent the real wind speed and direction observed at the 7 labelled stations. Evidently, the proposed model demonstrates a capability to capture local wind patterns in finer detail compared to the ECMWF forecast. Furthermore, the proposed model's performance is superior to the ECMWF forecast at most labelled stations.

\subsection{Cost-Benefit Analysis of Model}
\label{sec:result:cost}
We trained the model on an NVIDIA A100 40GB GPU, the test were conducted on a CPU machine. 
For training, the batch size was set to 64. We set the learning rate to 0.001, epochs to 200, and early stop patience to 5. 
The model training and testing cost is stated in Table \ref{tab:cost}. 

\begin{table}
\centering
\caption{\textbf{Model training and testing cost}}\label{tab:cost}
\resizebox{\linewidth}{!}{%
\begin{tblr}{
  vline{2} = {-}{},
  hline{2} = {-}{},
}
      & {Data size\\(space)} & {Data size\\(time)} & Time         & Memory  & Machine              \\
Train & 3.4$^\circ$ x 3.4$^\circ$    & 2 years             & 110 hours    & 17.3 GB & {GPU: NVIDIA A100\\40GB GPU} \\
Test  & 3.4$^\circ$ x 3.4$^\circ$    & 52 hours            & 5.57 minutes & 757 MB  & {CPU: 2 x AMD EPYC \\7313 4Core }                   
\end{tblr}}
\end{table}

The Python code used for training and testing the model, together with the plots of the outcome, can be found in the GitHub repository \url{https://github.com/FulingChen/Wind_s3_public.git}

\section{Conclusion and Future Work}
\label{sec:conclusion}
In this study, we have demonstrated the potential of our machine learning model for wind forecasting across various temporal horizons. Through a semi-supervised learning approach and a broader geographic perspective, our model shows promise in capturing transient spatial and temporal patterns, enabling accurate predictions from very short-term to medium-term, and potentially long-term forecasts. By addressing limitations related to data gaps and spatial inconsistencies, our model offers high-resolution gridded predictions over expansive areas, providing detailed insights crucial for applications like wind farm optimisation and bushfire management.

Furthermore, our study fills a notable gap in existing machine learning-based wind forecasting research by integrating a comprehensive suite of meteorological factors into the predictive model. Through the amalgamation of advanced techniques with factors such as terrain, mean sea level air pressure, humidity, and temperature, our approach offers a holistic representation of wind dynamics. These findings highlight the potential of machine learning in enhancing wind forecasting accuracy and reliability, with implications for sectors like renewable energy, agriculture, and disaster management. Continued research and refinement of predictive models will further advance our ability to sustainably harness wind resources and mitigate extreme weather events' impact.

\appendix
\section{\break ABED architecture}
\label{sec:app1}

\begin{figure}[tbp]
    \centering
    \includegraphics[width=0.8\textwidth]{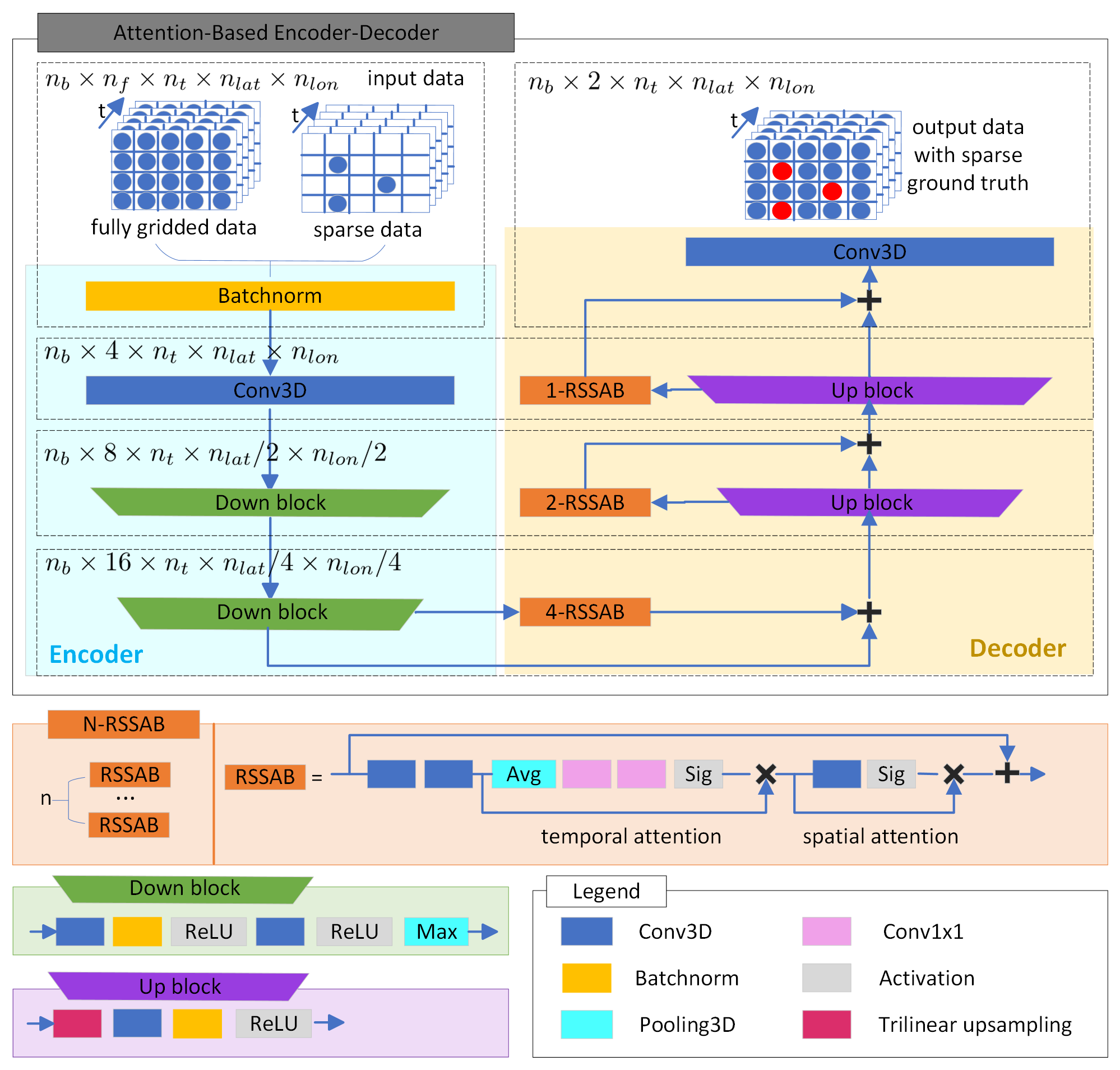} 
    \caption{The architecture of the ABED model.}
    \label{fig:model}
\end{figure}

The applied ABED model architecture is presented in Figure \ref{fig:model}

\section*{Acknowledgment}

We would like to acknowledge the Department of Primary Industry and Regional Development of Western Australia for granting access to Weather API 2.0, which provided essential weather observation data from their weather stations for public use, free of charge. Additionally, we are grateful to the European Centre for Medium-Range Weather Forecasts (ECMWF) for their publicly available archived reanalysis dataset through the Copernicus Climate Data Store (CDS). 

\bibliographystyle{IEEEtran}

\bibliography{access}

\end{document}